%% file: main.tex
\PassOptionsToPackage{table}{xcolor}
\documentclass[10pt,logo,copyright]{template/motiotechreport}
\usepackage{parskip}
\usepackage{longtable,ragged2e,float}
\usepackage[section]{placeins}
\usepackage[authoryear,sort&compress,round]{natbib}
\let\autocite\citep
\let\textcite\citet

\definecolor{ink}{HTML}{17232D}
\definecolor{resultviolet}{HTML}{8D91CF}
\definecolor{resultblue}{HTML}{73B6CC}
\definecolor{pastelviolet}{HTML}{D5D7EE}
\definecolor{pastelpink}{HTML}{F3D1D1}
\definecolor{pastelblue}{HTML}{D6EAF0}
\hypersetup{colorlinks=true,citecolor=ink,urlcolor=ink,linkcolor=ink,pdfauthor={SpatialtemporalAI},pdftitle={MOTIO: Agentic Skill Engineering for Reality-Grounded Humanoid Loco-Manipulation}}
\graphicspath{{./}}
\title{MOTIO: Agentic Skill Engineering for Reality-Grounded Humanoid Loco-Manipulation}
\author{\small SpatialtemporalAI}
\begin{document}
\maketitle
\noindent\includegraphics[width=\textwidth]{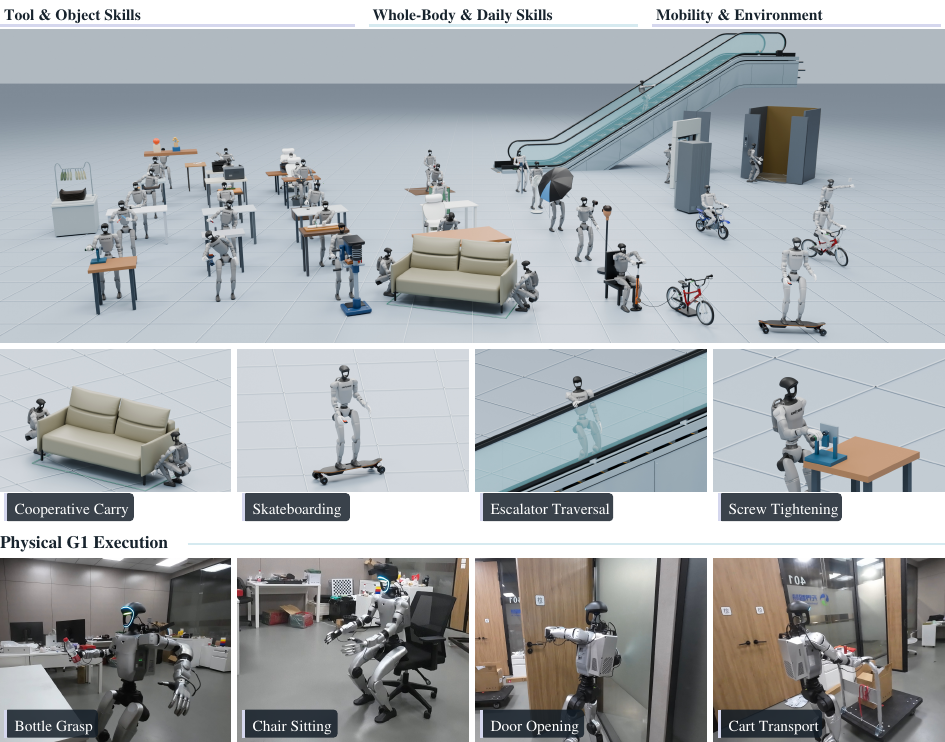}
\captionof{figure}{MOTIO develops diverse contact-rich humanoid loco-manipulation skills. Top: 36 human-reviewed skill states arranged in a shared virtual gallery, with tool and object skills (left), whole-body and daily skills (center), and mobility and environment interaction (right). Middle: cooperative carry, skateboarding, escalator traversal, and screw tightening. Bottom: physical G1 execution of bottle grasp, chair sitting, door opening, and cart transport.}
\label{fig:motio-results-teaser}

\clearpage
\input{main_content.tex}
\FloatBarrier
\bibliographystyle{plainnat}
\bibliography{references}
\input{authors.tex}
\end{document}


\maketitle
\input{supplement_content.tex}
\FloatBarrier
\bibliographystyle{plainnat}
\bibliography{references}

%% file: main_content.tex
\begin{abstract}
We present MOTIO, an agentic skill-engineering system that turns ordinary human-action videos, reality-grounded 3D assets, motion references, and reusable robot skills into executable humanoid loco-manipulation programs. Through in-context task and contact reasoning, an agent constructs staged robot-native interactions and improves them using physics execution, multi-frame feedback, and human validation---without task-specific teleoperation or robot demonstration collection. Agentic 3D supplies metric product and environment assets, while existing whole-body, inverse-kinematics, and dexterous-hand interfaces realize the programs. Across 33 timestamped first approvals, the median recorded agent-turn time is 97.0 minutes and request-to-approval duration is 2.80 hours. Simulation-side layout and object variation prepare a long-horizon bottle-disposal skill for a continuous 101.75-second physical G1 execution in a scanned office; physical refinement reduces screw-tightening thread residual by 76.0\%. Agent-generated skills additionally supervise BC/DAgger distillation into one shared reference policy, completing 18 simulated tasks under the same checkpoint without inference-time expert intervention.
\end{abstract}
\abscontent
\section{Introduction}\label{introduction}

Humanoid foundation controllers increasingly provide reusable walking, balancing, and whole-body tracking. Developing a new contact-rich task in a real environment additionally requires metric object geometry, interaction stages, and execution under robot dynamics. Human demonstration videos reveal the sequence and intent of contact events---reaching, grasping, turning, regrasping, releasing, and checking the terminal state---which can be translated into robot-space programs.

MOTIO provides an agentic task-acquisition interface. An Agentic AI harness receives a human-action video, a natural-language objective, metric scene or product assets, motion references, robot-specific skills, and prior executable cases. It reasons over task phases and contacts in context, constructs an editable robot-native program, and iterates on physical state and multi-frame visual feedback. Human review selects executable skill versions for reuse.

MOTIO has two stages. \textbf{Agentic 3D} builds metric product and scene assets, primarily in Blender and with FreeCAD for parameterized or standards-oriented geometry. \textbf{Agentic Skill Engineering} translates human-action context into contacts, goals, phase transitions, and robot references; MuJoCo \autocite{todorov2012mujoco} provides execution feedback for program revision. Measured room layouts, RGB-D observations, LiDAR localization, and existing whole-body and hand control enable selected skills to run on physical G1. Simulated skill executions also serve as expert references for one shared BC/DAgger policy, without new task-specific reinforcement-learning rewards.

Representative tasks show how the agent turns context into robot behavior. For a HEUER vise, human-action video and prior programs guide a 48-second right-hand grasp--turn--regrasp--clamp skill. For a Rapesco stapler, human feedback about palm pressing and withdrawal changes the contact and retreat phases. MOTIO spans locomotion, whole-body contact, articulated mechanisms, tool use, materials, and multi-humanoid tasks. In an office reconstructed with 3D Gaussian splatting \autocite{kerbl2023gaussiansplatting}, G1 completes bottle disposal and door/cart transport; agent-generated references also support a shared policy that executes 18 simulated tasks under one checkpoint.

Our contributions are threefold. \textbf{(1)} We introduce demonstration-conditioned \emph{in-context skill engineering}, constructing robot-native loco-manipulation programs from human-action video, reusable skills, and prior cases without collecting task-specific teleoperation or robot demonstrations. \textbf{(2)} We connect Agentic 3D assets, physics-grounded program revision, and human validation into a reusable task-development workflow. \textbf{(3)} We evaluate diverse humanoid tasks through development budgets, contact refinement, Real2Sim2Real execution, and distillation of agent-engineered skills into a shared multitask reference policy.

\section{Related Work}\label{related-work}

\textbf{Reusable humanoid motion and world-grounded generation.} SONIC
\autocite{luo2026sonic}, ScaleBFM \autocite{zeng2026scalebfm}, and
Kimodo \autocite{rempe2026kimodo} provide increasingly powerful motion
tracking or generation capabilities. GRAIL \autocite{xie2026grail}
generates world-conditioned human--object interactions, reconstructs
metric 4D interaction, retargets to humanoids, and trains task-general
trackers. Object manipulation uses an object-aware adaptor with a frozen
motion controller; scene interactions use adapted scene-aware tracking.
It further trains egocentric visual policies for robot deployment. MOTIO
uses reusable motion controllers within another task-acquisition route:
human-action video supplies phase and contact semantics to a coding
agent, which constructs and physically refines an editable program in
robot space.

\textbf{Agentic scene and program construction.} RoboSimStudio
\autocite{chen2026robosimstudio}, Real2Gym \autocite{ren2026real2gym},
Agentic RSR \autocite{li2026agenticrsr}, and PhysEvo
\autocite{tian2026physevo} highlight editable interactive scenes,
executable task tools, and feedback-directed improvement. USDCraft \autocite{zhang2026usdcraft} writes and revises executable programs from partial metric geometry, rechecking geometry while authoring articulated USD simulation assets. MOTIO connects these scene-authoring ideas to long-horizon whole-body loco-manipulation through explicit staged contact programs and shared embodiment interfaces.

\textbf{Human-action learning and retargeting.} GPT-Policy
\autocite{cheng2026icl} and RoboICL \autocite{liu2026roboicl} motivate
using visual task context and prior experience to guide embodied
execution. A GPT-6-Astra study
\autocite{sensenova2026gpt6astraachievedembodiedintelligence} examines
demonstration-derived in-context packages and harness designs on
embodied benchmarks, alongside zero-shot deployment across eight
real-robot configurations. Contact-aware methods such as HOI-Retarget
\autocite{shin2026hoiretarget}, VICAR \autocite{kalaria2026vicar},
OmniRetarget \autocite{yang2026omniretarget}, and InterMimicGen
\autocite{zhang2026intermimicgen} reconstruct or retarget detailed
human--object motion. MOTIO instead treats video as a source of task and
contact \emph{meaning} for robot-native program synthesis. The resulting
program can be tested directly in physics and reused as an expert data
source.

\section{Agentic Skill Engineering}\label{agentic-skill-engineering}

\begin{figure}[!htbp]
\centering
\includegraphics[width=\linewidth]{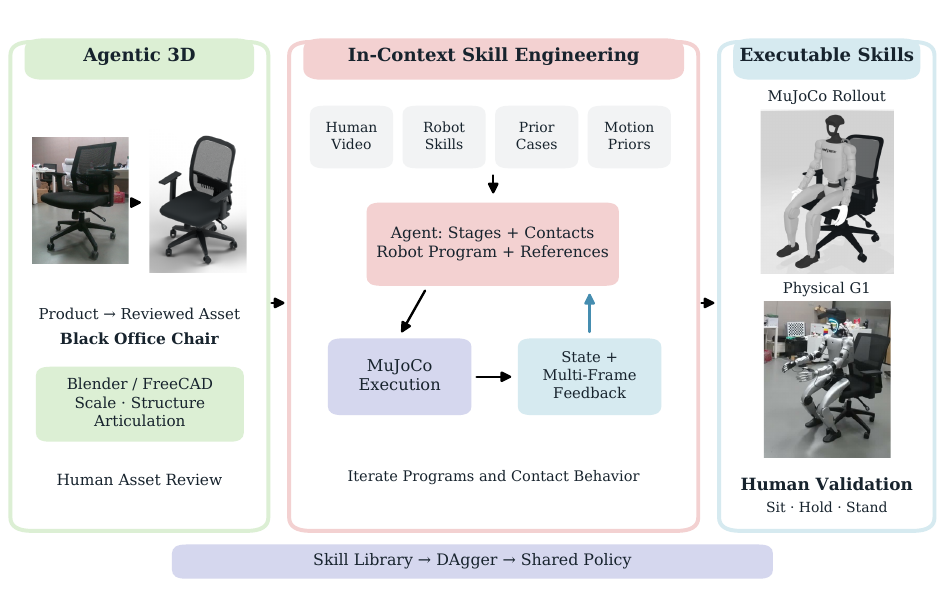}
\caption{MOTIO workflow illustrated with G1 sitting on a black office chair. A reviewed asset enters video- and skill-conditioned program synthesis, followed by MuJoCo execution, multi-frame feedback, and human validation with physical G1 execution. Skill-library executions support downstream shared-policy learning.}
\label{fig:motio-overview}
\end{figure}

\subsection{Reality-grounded assets and the executable task
artifact}\label{reality-grounded-assets-and-the-executable-task-artifact}

\begin{figure}[htbp]
\centering
\includegraphics[width=0.94\linewidth]{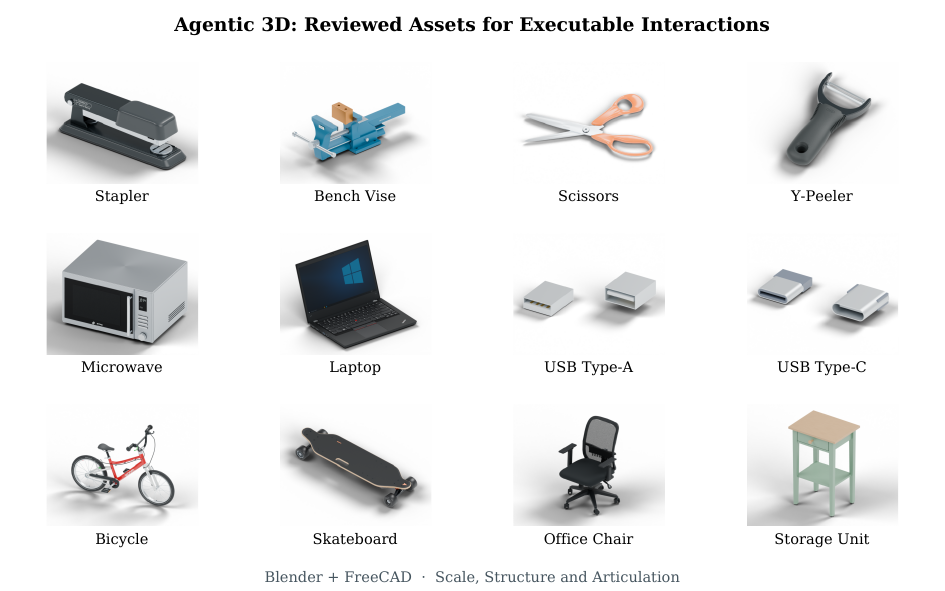}
\caption{Representative Agentic 3D assets from everyday tools, engineered interfaces and office objects. Metric geometry and component structure support task-specific physical interaction models (Supplementary Section S1).}
\label{fig:asset-gallery}
\end{figure}

Stage I takes a natural-language product or scene specification, gathers dimensions and reference material, and constructs a metric asset for human review. Stage II receives the task objective, a human-action demonstration, the reviewed asset, an interactive physical environment, and the robot embodiment and controllers. The agent then produces robot-frame targets and event-driven phase transitions. The asset catalogue contains 78 local families and 104 simulation-task families, with 59 shared families and 123 distinct families in their union; these units cover products, supporting objects, and scenes.

Making an asset executable requires collision geometry, articulated bodies and joints, named contact sites, and task-event measurements. The 159 \ensuremath{\times} 41 \ensuremath{\times} 53 mm Rapesco stapler has lever, magazine, anvil, and press-cap representations; the robot uses stapler-frame palm targets and paper-frame stabilization targets while the event logic checks lever motion and sustained contact. The HEUER vise couples a translating jaw and rotating screw so the robot's palm arc produces measurable clamping. Supplementary Section S1 specifies these physical interfaces.

\subsection{Video-conditioned program
synthesis}\label{video-conditioned-program-synthesis}

\begin{figure}[htbp]
\centering
\includegraphics[width=\linewidth]{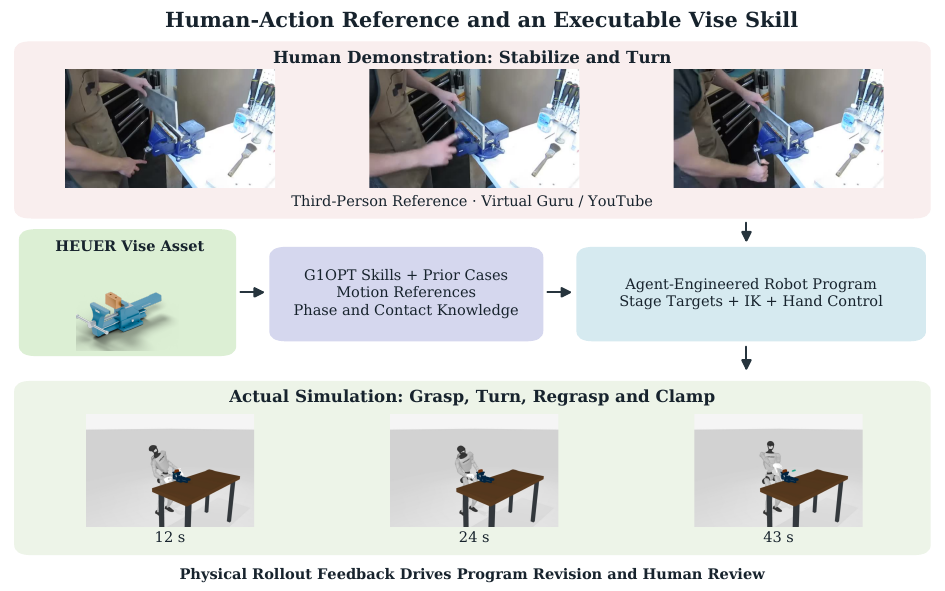}
\caption{Video-conditioned task engineering for a HEUER vise. Top: an illustrative third-person vise operation; middle: asset and robot-skill context; bottom: G1 MuJoCo grasp, turn, regrasp and clamp at 12, 24 and 43 s, followed by handle release and workpiece retention.}
\label{fig:video-icl}
\end{figure}

The task agent brings human-action frames, the objective, metric assets, motion references, G1/Revo2 skill instructions, prior programs, and embodiment constraints into context. Video establishes the action sequence and contact intent; object and robot models provide scale, coordinate frames, reachable targets, and event thresholds. The agent writes body and hand references with phase transitions that respond to simulated state.

In the HEUER vise task, human-video cues for stabilization, rotation, regrasping, and release become robot-specific phases: initial workpiece support, a right-hand palm arc around the handle, hand retreat and repositioning, and finally withdrawal followed by a retention check. The 48-second program completes four seven-second half-turn cycles, producing 2.052 handle turns and reducing jaw opening from 38.000 to 29.987 mm. The workpiece remains clamped after hand and support withdrawal. Figure~\ref{fig:video-icl} illustrates the phase-to-program correspondence and G1 execution; Supplementary Section S2 details the video context and program stages.

\subsection{Physics-grounded revision and human
approval}\label{physics-grounded-revision-and-human-approval}

For each candidate program, the agent runs MuJoCo, inspects robot and object states alongside a multi-frame contact sheet, and revises targets, phase timing, contact logic, or control references. Human feedback specifies task intent and motion quality; state-based checks determine whether the desired physical event occurred.

For the Rapesco stapler, human feedback requests palm-driven pressing and a more natural hand withdrawal. The program is revised into palm press, unload, paper release, elbow retraction, and arm lowering. The resulting 31-second execution triggers the contact-gated staple event at 14.8125 s, leaves the paper joined, and withdraws both hands. The skill receives human approval. The screw-tightening task provides a second example: simulation feedback reduces maximum thread residual from 269.44 to 64.58 \ensuremath{\mu}m. Supplementary Section S3 provides the program phases and event checks.

\subsection{Embodied deployment and optional policy
reuse}\label{embodied-deployment-and-optional-policy-reuse}

Frozen ScaleBFM whole-body tracking executes standing and locomotion references, with task-specific inverse kinematics, Revo2 finger schedules, and bounded hand/contact feedback. MuJoCo is the main validation engine; Isaac Sim/PhysX models deformable fruit peel in the mango-peeling task. For physical bottle disposal, Livox MID-360 localization, RGB-D and SAM3 observations, and a measured bin pose guide Thor's staged planning. The resulting body and hand references are validated in MuJoCo before execution on G1.

Skill programs and their simulated executions provide expert state/reference pairs for a task-conditioned BC/DAgger student. The student predicts six 59-dimensional references at 20-ms intervals, executed through frozen ScaleBFM and the existing contact/IK feedback stack. DAgger labels student-visited states with the skill-program expert and uses task-event gates for recovery. The final reference policy conditions on current task and phase state plus explicit geometry memory (\texttt{reference\_history=False}); the observation channels and training interface are detailed in Supplementary Section S6.

\section{Reality-Grounded
Loco-Manipulation}\label{reality-grounded-loco-manipulation}

\subsection{From a scanned office to an interactive
task}\label{from-a-scanned-office-to-an-interactive-task}

\begin{figure}[htbp]
\centering
\includegraphics[width=\linewidth]{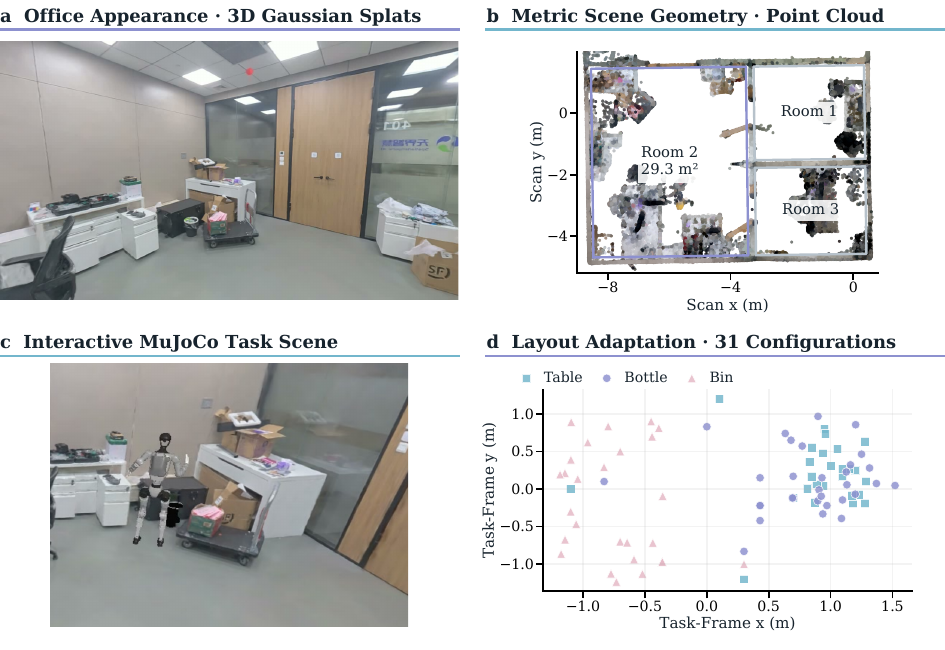}
\caption{Reality-grounded office reconstruction and task preparation. (a) Room 2 appearance rendered from the scanned 3D Gaussian model. (b) Metric scan geometry across three connected rooms; Room 2 spans 29.3 m$^2$. (c) The interactive MuJoCo bottle-to-bin scene. (d) Table, bottle and bin locations across 31 executed layout configurations.}
\label{fig:office}
\end{figure}

The reconstructed office has three connected rooms, with the bottle-disposal task located in Room 2 (29.3 m\ensuremath{^2}). Its 3DGS appearance model, metric point cloud, and collision/navigation representations share the task coordinate frame used for object measurements and localization. Figure~\ref{fig:office} shows the room geometry, interactive MuJoCo scene, and 31 tested layout configurations.

The bottle-to-bin skill includes grasp, turn, loaded walking, bin approach, release, and retreat. Simulation preparation covers three units of variation. \textbf{Thirty-one room layouts} complete full grasp--carry--release sequences with continuous carrying grasp (61.2--138.6 s). \textbf{Forty-four bottle-property settings}, spanning diameter 58--68 mm, height 220--235 mm, and mass 20--600 g, each pass the opposing-finger hand-fixture grasp window at 8\textless t\textless10 s followed by release. On twelve matched settings, full-window retention improves from 11/12 to 12/12 after contact refinement. \textbf{Three perception-conditioned whole-body variants} complete all eight stages, including carried-object grasp, release, and return to stand. Table~\ref{tab:sim2real-results} summarizes these outcomes; the distributions appear in Supplementary Section S4 and Figure S2.

\begin{figure}[htbp]
\centering
\includegraphics[width=\linewidth]{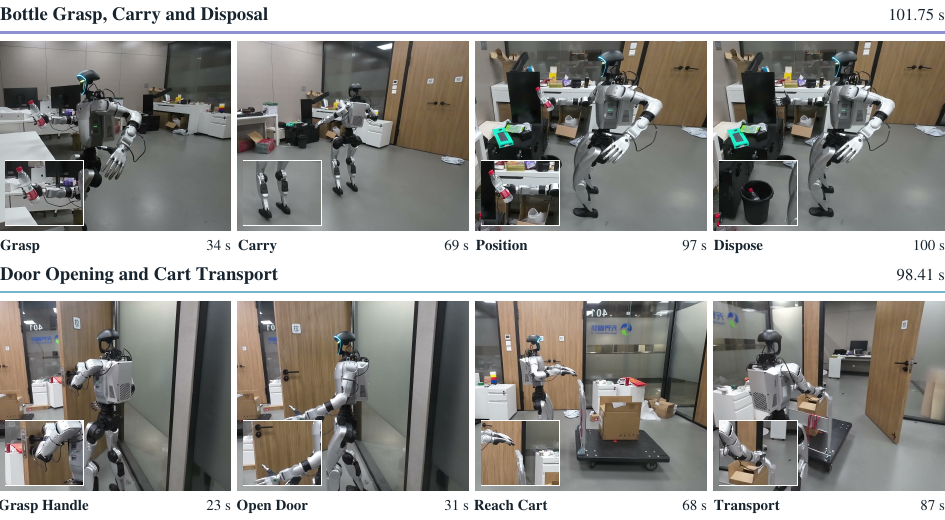}
\caption{Physical G1 deployment. Top: selected moments from the continuous 101.75-second bottle grasp, carry and disposal episode. Bottom: door opening followed by cart transport in a 98.41-second episode. Insets magnify contact and task details from the corresponding frames.}
\label{fig:bottle-preparation}
\end{figure}

For physical execution, a measured start pose, observed table and bottle, and bin location determine the staged task plan, which is validated in MuJoCo. G1 completes the full bottle pickup, turn, loaded walk, bin release, and retreat in one continuous \textbf{101.7543-second} execution without retry. A separate \textbf{98.41-second} G1 sequence opens a door and transports a cart. Figure~\ref{fig:bottle-preparation} shows phase-matched moments and contact details from these two tasks. Chair sitting, wall-switch interaction, and mouse drawing extend the physical repertoire; simulation tasks further cover balance, moving supports, collaboration, tools, and materials. Supplementary Section S7 lists the six physical demonstrations.

\input{deployment_results.tex}
\begin{figure}[!p]
\centering
\includegraphics[width=\linewidth]{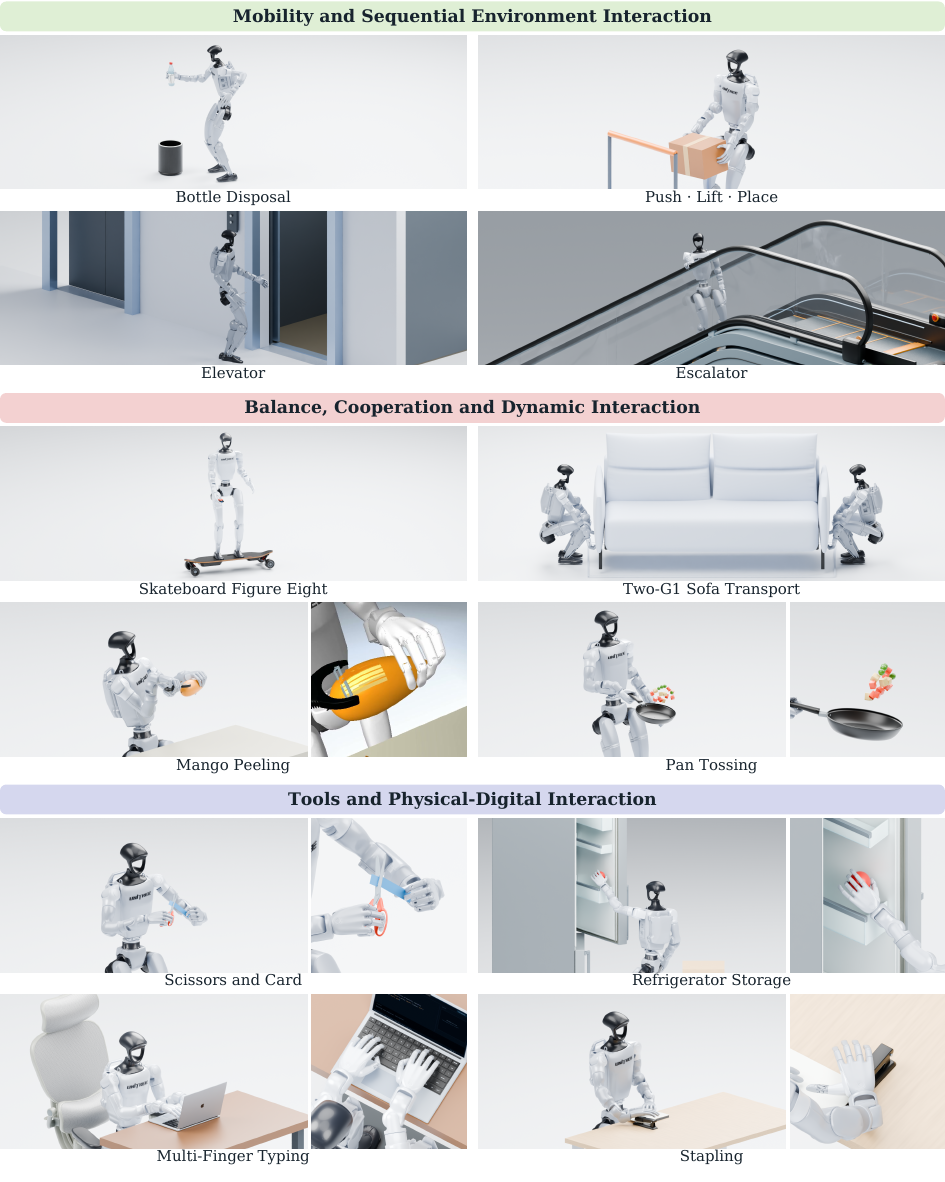}
\caption{Diverse humanoid skills spanning long-horizon mobility, environment interaction, balance, multi-robot cooperation, dynamic object handling, articulated tools and physical--digital interaction. Close-ups highlight task-critical hand--object interactions.}
\label{fig:diversity}
\end{figure}

\section{Quantitative Skill
Development}\label{quantitative-skill-development}

\subsection{Development time and execution-feedback
budget}\label{development-time-and-execution-feedback-budget}

\begin{figure}[htbp]
\centering
\includegraphics[width=\linewidth]{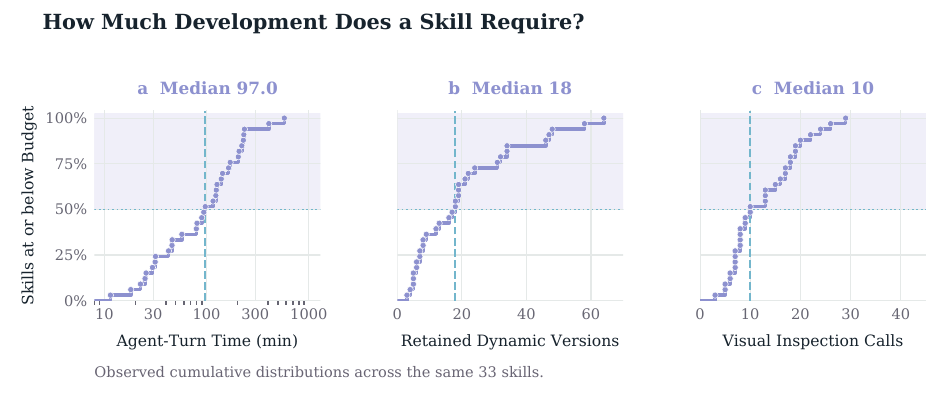}
\caption{Development budget across 33 skills with matched first human approval: empirical cumulative distributions for agent-turn time, retained dynamic trajectories and visual inspection calls. Dashed vertical lines mark medians.}
\label{fig:budget-ecdf}
\end{figure}

We study 34 consecutively registered skills (Cases 40--73), including 33 with task-request and first human-approval timestamps. \textbf{Agent-turn time} sums nonoverlapping open-turn intervals, including tool and resource waits; \textbf{request-to-approval time} spans the full calendar interval, including breaks and human review. Median first-approval time is \textbf{97.0 agent-turn minutes} (IQR 42.5--170.3) and \textbf{2.80 calendar hours} (IQR 1.33--9.80). The corresponding medians are \textbf{two development turns}, \textbf{18 retained dynamic versions}, \textbf{ten visual inspections}, and \textbf{one substantive human revision prompt}.

Within the 33-skill first-approval cohort, \textbf{25/33 (75.8\%)} reach approval within three recorded agent-turn hours, \textbf{21/33 (63.6\%)} within 20 retained dynamic versions, and \textbf{17/33 (51.5\%)} within ten visual-inspection calls. Figure~\ref{fig:budget-ecdf} shows the empirical budget distributions. Supplementary Section S5 provides per-skill time profiles and three examples in which later human requests expand sofa-carrying, skateboard, and flexible-bag skills.

\begin{table}[htbp]
\centering\small
\caption{Development budget to first human approval across 33 skills (median and interquartile range).}\label{tab:key-budget}
\renewcommand{\arraystretch}{1.14}
\begin{tabularx}{\linewidth}{@{}Xrr@{}}\toprule
\rowcolor{pastelviolet}\textbf{Measure} & \textbf{Median} & \textbf{IQR} \\ \midrule
Agent-turn (min) & 97.0 & 42.5--170.3 \\
Request-to-approval (h) & 2.80 & 1.33--9.80 \\
Agent turns & 2 & 1--2 \\
Retained dynamic versions & 18 & 7--31 \\
Visual inspections & 10 & 7--18 \\
Human revision prompts & 1 & 0--1 \\ \bottomrule
\end{tabularx}\end{table}

\subsection{Measurable physical
refinement}\label{measurable-physical-refinement}

\begin{figure}[htbp]
\centering
\includegraphics[width=\linewidth]{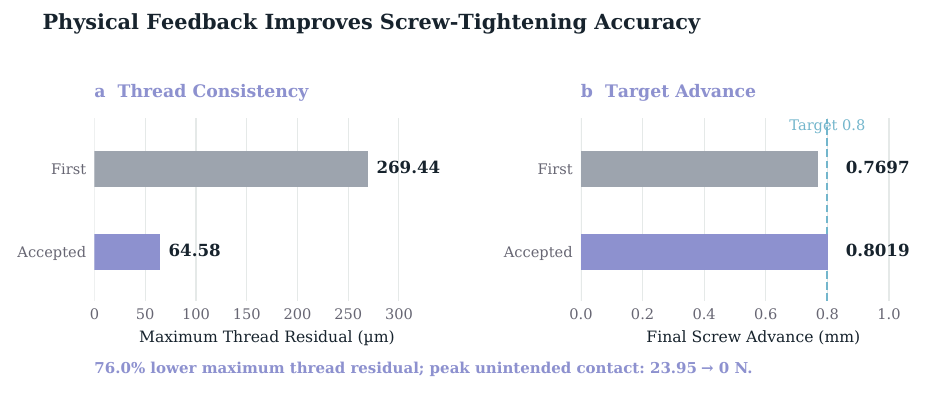}
\caption{Physical refinement of screw tightening from first to human-approved final candidate: maximum thread residual 269.44 to 64.58~$\mu$m (76.0\% lower) and final advancement 0.7697 to 0.8019 mm against 0.8 mm target.}
\label{fig:screw-outcome}
\end{figure}

Simulation feedback improves measurable contact outcomes. For screw tightening, maximum thread residual falls from \textbf{269.44 to 64.58 \ensuremath{\mu}m} (a \textbf{76.0\% reduction}), final screw advance moves from \textbf{0.7697 to 0.8019 mm} against a \textbf{0.8 mm} target, and peak unintended contact decreases from \textbf{23.95 N to 0 N}. Figure~\ref{fig:screw-outcome} compares the first and human-approved candidates; Supplementary Section S5 reports the full \texttt{full01}--\texttt{full04} sequence.

\section{One Shared Multitask Reference
Policy}\label{one-shared-multitask-reference-policy}

MOTIO skill programs serve both as executable task controllers and as experts for a shared reference policy. Behavior cloning initializes the student, and DAgger \autocite{ross2011dagger} adds expert labels on student-visited states. The \textbf{3,127,287-parameter} network outputs \textbf{6 \ensuremath{\times} 59} references to frozen ScaleBFM and the task-feedback executor. The ten-task extension dataset contains \textbf{63 episodes and 92,130 frames}, including 77,130 student-execution frames and 142 recovery-flag frames within that dataset.

\begin{figure}[htbp]
\centering
\includegraphics[width=\linewidth]{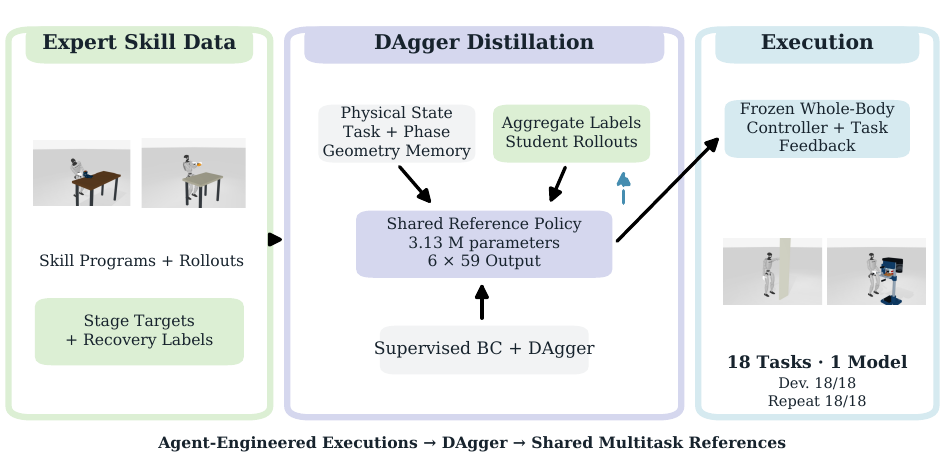}
\caption{Agent-engineered simulated skill executions provide BC initialization and DAgger expert references for a shared reference generator, executed through the frozen ScaleBFM whole-body controller and retained task-feedback interfaces.}
\label{fig:dagger}
\end{figure}

\begin{figure}[htbp]
\centering
\includegraphics[width=\linewidth]{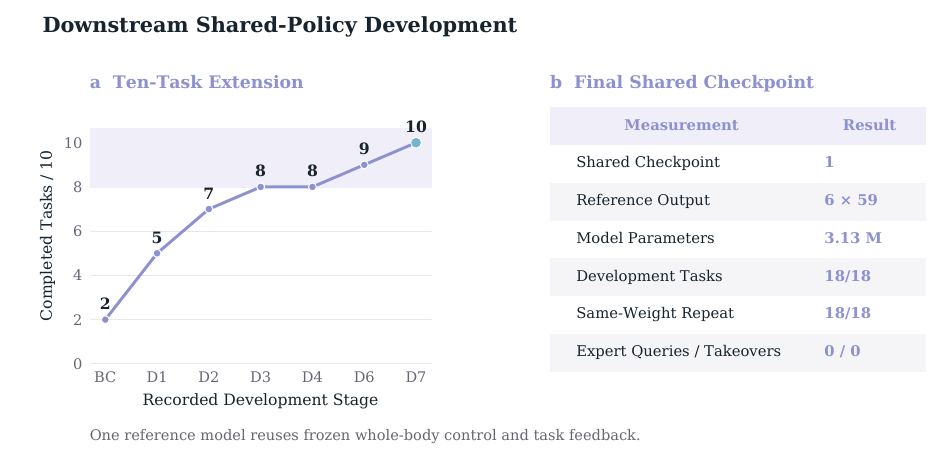}
\caption{Shared-policy development across ten added tasks (BC 2/10 to D7 10/10) and final one-checkpoint eighteen-task performance (18/18 development and 18/18 same-condition repeat; zero inference-time expert queries/takeovers). The stage trace includes program, representation and execution-interface refinements.}
\label{fig:shared-results}
\end{figure}

On ten added task types, completion progresses from \textbf{2/10 with BC} to \textbf{5/10 (D1), 7/10 (D2), 8/10 (D3 and D4), 9/10 (D6), and 10/10 (D7)} as aggregation, reference prediction, representation, and execution interfaces are developed. One shared checkpoint completes \textbf{18/18 simulated development tasks} and \textbf{18/18 same-condition repetitions}, with \textbf{zero inference-time expert queries or takeovers}. The 18 development runs total 673 simulated seconds (16--66 s per task). Agent-engineered skills thus provide reusable expert targets without task-specific reward engineering or additional task RL. Figures~\ref{fig:dagger} and~\ref{fig:shared-results} summarize the pipeline and evaluation; Supplementary Section S6 describes the interface and task suite.

\FloatBarrier
\section{Conclusion and Outlook}\label{conclusion-and-outlook}

MOTIO uses human-action context, metric assets, and existing control capabilities to engineer executable humanoid loco-manipulation skills. Agent-driven program synthesis and physics-grounded revision yield skills for contact-rich simulation tasks and selected real G1 executions, with human validation defining task acceptance. These same programs supply expert trajectories for a shared multitask reference policy. Parallel agent development, consistent physical verification, and accumulated skill libraries offer a path to broader task coverage.

%% file: deployment_results.tex
\begin{table}[htbp]
\centering
\caption{Simulation preparation and physical G1 execution.}
\label{tab:sim2real-results}
\small
\setlength{\tabcolsep}{4pt}
\renewcommand{\arraystretch}{1.12}
\begin{tabularx}{\linewidth}{@{}p{0.31\linewidth}X p{0.32\linewidth}@{}}
\toprule
\textbf{Study} & \textbf{Variation or sequence} & \textbf{Observed result} \\
\midrule
\multicolumn{3}{@{}l}{\textit{Simulation preparation}} \\
Full-task layouts & 31 configurations & 31/31 complete sequences \\
Bottle properties & 44 unique settings & 44/44 full-window grasps \\
Paired grasp refinement & Same 12 settings & 11/12 $\rightarrow$ 12/12 full-window retention \\
Perception-conditioned execution & 3 whole-body variants & 3 complete eight-stage sequences \\
\midrule
\multicolumn{3}{@{}l}{\textit{Physical G1 execution}} \\
Bottle pickup, carry and disposal & Continuous episode & 101.75 s \\
Door opening and cart transport & Continuous episode & 98.41 s \\
\bottomrule
\end{tabularx}
\par\smallskip
{\footnotesize Grasp-window retention is measured over 8--10 s; each whole-body variant executes all eight stages.}
\end{table}

%% file: authors.tex
\section*{Authors and affiliations}
\begingroup
\small
\noindent
Shijian Ma\textsuperscript{1,*},
Zhe Li\textsuperscript{1,2,*,\textdagger},
Shang Gao\textsuperscript{1,3,*},
Junling Tu\textsuperscript{1,2},
Binzhang Hu\textsuperscript{1},
Feng Zheng\textsuperscript{1,2}.

\noindent\textsuperscript{1}Spatialtemporal AI\\
\textsuperscript{2}Southern University of Science and Technology\\
\textsuperscript{3}University of Macau

\noindent\textsuperscript{*}Equal contribution (co-first authors).\\
\textsuperscript{\textdagger}Project Lead: Zhe Li.
\endgroup

%% file: supplement_content.tex
\section{Agentic 3D assets and task-executable physical
representations}\label{agentic-3d-assets-and-task-executable-physical-representations}

\subsection{Asset production and library
units}\label{asset-production-and-library-units}

Agentic 3D starts from a product or environment specification, searches product references and manufacturer dimensions, and constructs an editable model with metric scale and component structure. Blender handles most geometry, while FreeCAD supports parametric constraints, drawings, and engineered interfaces. Human asset review selects the model used to build contact and dynamics representations. Visual meshes, transforms, component names, and frames remain available when the interaction model is created.

Asset counts use \emph{families} as the unit. The local msj-3d library contains \textbf{78 families} (73 product/paired kits, 3 scenes, and 2 hardware assemblies); simulation tasks use \textbf{104 families}; \textbf{59 families} overlap, yielding \textbf{123 distinct families}. Historical revisions, repeated instances, and component meshes are grouped within their families. The local library comprises 71 Blender-only families, five FreeCAD-only families, and two represented in both formats.

\begin{figure}[htbp]
\centering
\includegraphics[width=\linewidth]{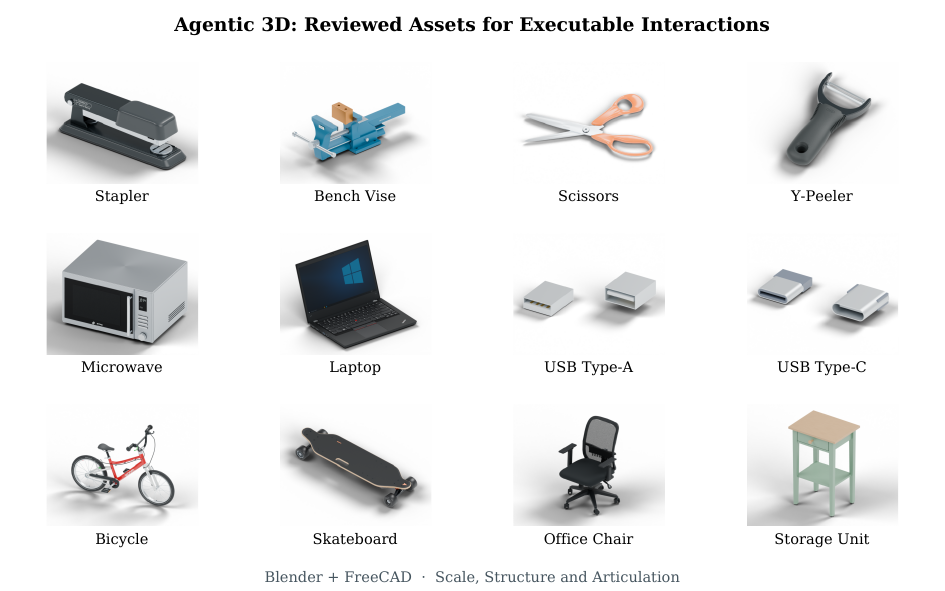}
\caption{Agentic 3D families across tools, engineered interfaces, product mechanisms and office furnishings.}
\label{supp:fig:assets}
\end{figure}

\subsection{From reviewed product to physical contact
model}\label{from-reviewed-product-to-physical-contact-model}

A reviewed asset supplies visible geometry, dimensions, and component hierarchy. The task-ready physical model assigns collision proxies, articulated joints, passive constraints, contact sites, and event measurements. The agent addresses named object frames to generate reachable robot targets. Human asset review checks geometry and structure; human skill approval assesses the executed outcome and motion quality.

\textbf{Rapesco stapler.} The \texttt{Rapesco\_Marlin\_Stapler.blend} model measures approximately \textbf{159 \ensuremath{\times} 41 \ensuremath{\times} 53 mm}. Visible base, lever, and magazine meshes follow rigid bodies, while collision proxies represent the press cap, driver, anvil, and paper support. Lever and magazine rotate about local y axes against task-model springs (lever range 0--0.125 rad, spring 18; magazine range 0--0.075 rad, spring 1.5). A one-way lost-travel mechanism couples driver motion to lever rotation. Three supported A5 sheets define the initial workpiece. The staple event requires the driver/press contact configuration for \textbf{12 steps at 400 Hz (30 ms)}, including right-palm loading and left-hand paper stabilization; it activates a paper connection proxy. Press-cap, driver, and paper frames provide the palm and supporting-hand targets.

\textbf{HEUER vise.} HEUER 100 product references specify a 100 mm jaw width, 125 mm maximum opening, and 50 mm clamping depth. The physical model contains a fixed seat, sliding jaw, screw, translating handle, and initially supported workpiece. The jaw moves along a local x axis and couples to screw rotation at a \textbf{4 mm/turn task-model lead}; the handle translates along local y within \ensuremath{\pm}72 mm. Screw angle determines the right-palm arc and regrasp targets. Task checks monitor jaw opening, opposing clamp contacts, workpiece motion, and retention following withdrawal.

\begin{table}[htbp]\centering\small
\caption{Task-facing asset to physics interface examples.}\label{supp:tab:asset}
\begin{tabularx}{\linewidth}{@{}p{27mm}XX@{}}\toprule
\rowcolor{pastelviolet}\textbf{Product asset} & \textbf{Physical model} & \textbf{Skill interface / event} \\ \midrule
Rapesco 159$\times$41$\times$53 mm & Base, lever, magazine, press cap, driver, anvil and contact proxies; articulated lever and magazine. & Right-palm press-cap target, left-hand paper stabilization, contact-gated paper connection and withdrawal. \\
HEUER vise & Fixed seat, translating jaw, screw, moving handle, supported workpiece; screw/jaw coupled at model lead 4 mm/turn. & Hand arc referenced to screw angle; opposing clamp, jaw clearance and retained workpiece measured at task completion. \\ \bottomrule
\end{tabularx}\end{table}

\section{Human demonstration as staged contact
context}\label{human-demonstration-as-staged-contact-context}

\subsection{Recovered reference
observations}\label{recovered-reference-observations}

The vise task begins with a natural-language request to clamp a workpiece by turning the handle and to consult human operation video. The agent views a steel-vise operation clip and an Ono Kosuki/Pexels handwheel clip, using their stabilization, rotation, regrasp, and release events to plan contact stages. The Pexels clip depicts workpiece support at approximately 0.5 s and hand repositioning near 2.5 and 4.8 s. The HEUER object model supplies geometry and reachable robot targets. Main Figure~4 uses an illustrative third-person vise demonstration to visualize these phases.

\subsection{Accepted Case 43 staged
program}\label{accepted-case-43-staged-program}

The 48-second HEUER skill starts with right-hand reach (0--6 s) and handle closure (6--8 s), followed by four seven-second half-turn cycles from 8--36 s. Each cycle contains turn (0--3.5 s), unflex (3.5--4.4 s), retreat (4.4--5.3 s), reposition (5.3--6.3 s), and regrasp (6.3--7.0 s), with three intermediate regrips. The goal function computes wrist/palm arcs around the handle from screw angle and maintains clearance during regrasp. Final tightening occurs at 36--37 s, right-hand release at 37--39 s, workpiece-support withdrawal at 39--41 s, and retention checking at 41--48 s. The skill produces \textbf{2.052235 handle turns}, reduces jaw opening from \textbf{38.000 to 29.987218 mm}, and retains the workpiece after withdrawal.

Table~\ref{supp:tab:vise} summarizes the accepted right-hand stages and physical checks.

\subsection{Additional video-guided tool
example}\label{additional-video-guided-tool-example}

A second tool example uses human references from Helpful DIY (ratcheting screwdriver at approximately 1:20) and Sikana (screwdriver operation at approximately 1:55--2:05). These establish axis support, wrap grasp, ratcheting rotation, and axial pressure. The robot program realizes those interactions around a supported horizontal tool axis and an M5 \ensuremath{\times} 0.8 screw mechanism.

\begin{table}[htbp]\centering\small
\caption{HEUER vise right-hand stages and physical checks.}\label{supp:tab:vise}
\begin{tabularx}{\linewidth}{@{}p{22mm}X X@{}}\toprule
\rowcolor{pastelviolet}\textbf{Window} & \textbf{Program phase} & \textbf{Observation} \\ \midrule
0--8 s & Reach and grasp handle bar; workpiece initially supported. & Initial reach/contact. \\
8--36 s & Four 7-s half-turn cycles with unflex, retreat, reposition, regrasp. & Rotation, contact and jaw travel. \\
36--41 s & Final tighten, release hand and withdraw support. & Clamp loading and release. \\
41--48 s & Hold test after both hand and support withdraw. & Workpiece retention and gap (38 to 29.987 mm). \\ \bottomrule
\end{tabularx}\end{table}

\section{Execution-grounded skill revision and control
implementation}\label{execution-grounded-skill-revision-and-control-implementation}

\subsection{Stapler program changes following human
feedback}\label{stapler-program-changes-following-human-feedback}

Human feedback on the stapler requests a palm-driven press and natural hand withdrawal. The revised program uses \texttt{palm\_press} (13--16 s), \texttt{hold\_stroke} (16--17 s), \texttt{unload\_palm} (17--19 s), \texttt{release\_papers} (19--21 s), \texttt{elbows\_retract} (21--24 s), \texttt{lower\_arms} (24--29 s), and final completion (29--31 s). The right palm aligns with the press cap while the left hand stabilizes paper. Lever-angle error around a 0.108-rad target adjusts palm height; after the staple event, the palm holds briefly before unloading and retreating with tabletop clearance.

During the 31-second physics rollout, the \texttt{clinched\_staple\_proxy} event occurs at \textbf{14.8125 s}, with approximately \textbf{14.8893 N} between the right palm pad and press cap and contact maintained by the supporting left hand. The event requires \textbf{30 ms} of sustained contact conditions. The mechanism returns passively, the paper remains connected on the table, and both hands withdraw. The completed skill receives human approval.

\begin{table}[htbp]\centering\small
\caption{Rapesco feedback-driven program change.}\label{supp:tab:revision}
\begin{tabularx}{\linewidth}{@{}p{29mm}XX@{}}\toprule
\rowcolor{pastelviolet}\textbf{Review input} & \textbf{Program revision} & \textbf{Physical result} \\ \midrule
Palm-driven press and natural hand withdrawal. & 13--16 s palm press, 16--17 hold, 17--19 unload, 19--21 paper release, 21--24 elbow retract, 24--29 lower arms. & Staple connection event 14.8125 s; paper joined, hands withdrawn by 31 s; human approval. \\ \bottomrule
\end{tabularx}\end{table}

\subsection{Physical execution and specialized material
simulation}\label{physical-execution-and-specialized-material-simulation}

MuJoCo \autocite{todorov2012mujoco} handles rigid-body and articulated-object dynamics. Standing and locomotion references use frozen ScaleBFM with task-specific IK, Revo2 hand targets, and bounded contact feedback. Supported or seated precision skills can use gravity-compensated joint control and local IK. For mango peeling, the G1 trajectory from MuJoCo drives a deformable fruit-peel representation in Isaac Sim 5.1 PhysX. LeHome-based tools provide additional deformable simulation capabilities.

\section{Office reconstruction, layout variation and
Real2Sim2Real}\label{office-reconstruction-layout-variation-and-real2sim2real}

\subsection{Metric scene and
localization}\label{metric-scene-and-localization}

The office reconstruction uses 3D Gaussian splatting \autocite{kerbl2023gaussiansplatting} across three connected rooms; Room 2 covers \textbf{29.3 m\ensuremath{^2}}. The scene includes approximately \textbf{1.86 million Gaussians}, a sparse point cloud of \textbf{1.17 million colored points}, and collision, navigation, and interaction representations. Task and scan frames align to measured room geometry. A G1-mounted Livox MID-360 localizes the robot; head D435i RGB-D and external SAM3 estimate the bottle and table, while the bin position comes from room measurements. Thor plans body and Revo2 hand references, validates the task in MuJoCo, and executes through ScaleBFM M Mode 7 with absolute-position feedback.

\subsection{Full-task layouts (31
runs)}\label{full-task-layouts-31-runs}

The 31 layouts comprise eleven named configurations (eight single-factor changes and three rear/left/right table placements), twelve development-random layouts, and eight random layouts generated after controller freezing. The latter use seed \texttt{2026092301}. Task-frame sampling bounds are: table (x,y) \ensuremath{\in} {[}0.7,1.4{]} \ensuremath{\times} {[}-0.25,0.75{]} m; bottle relative to table \ensuremath{\in} {[}-0.27,0.27{]} \ensuremath{\times} {[}-0.37,0.37{]} m; bin \ensuremath{\in} {[}-1.2,-0.25{]} \ensuremath{\times} {[}-1.3,1.0{]} m. Candidates with table/bin overlap, insufficient tabletop support, or infeasible stance/path are rejected. Each layout regenerates stance, approach angle, avoidance route, and phased body/hand targets. All \textbf{31/31} complete the grasp--carry--release task with continuous carrying grasp and return to stand in \textbf{61.2--138.6 s}.

\subsection{Hand-fixture object properties and paired
revision}\label{hand-fixture-object-properties-and-paired-revision}

The \textbf{44 simulated bottle-property settings} vary diameter \textbf{58--68 mm}, height \textbf{220--235 mm}, and mass \textbf{20--600 g}, with geometry, mass, and inertia updated together. They include 27 grid points, twelve Latin-hypercube samples, and five additional unique settings. The LHS samples use $n=12$, seed \texttt{20260923}, and randomized offsets within twelve strata before mapping to the parameter ranges. A hand fixture checks opposing thumb-plus-two-finger grasp contact during \textbf{8\textless t\textless10 s}, followed by release. All \textbf{44/44} settings pass this contact window.

\begin{figure}[htbp]
\centering
\includegraphics[width=\linewidth]{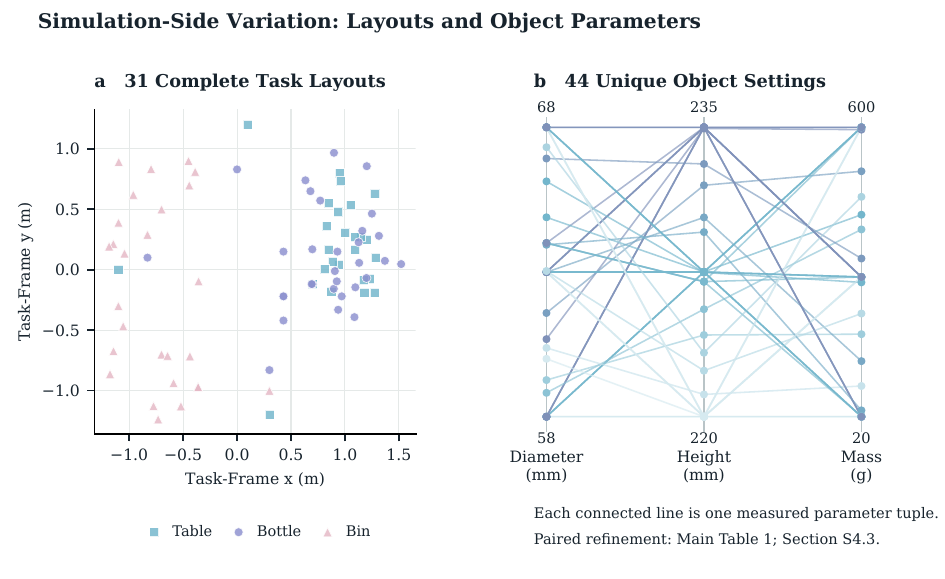}
\caption{Detailed simulation-side variation. Left: positions of table, bottle and bin for 31 full-task layouts. Right: 44 unique simulated bottle-parameter settings across diameter, height and mass; each connected line represents one tested setting.}
\label{supp:fig:coverage}
\end{figure}

Contact refinement is measured on the \textbf{same twelve LHS settings} within the 44. The earlier configuration retains contact throughout the window in 11/12 settings (mean 95.83\% contact-window retention); the revised version retains contact in 12/12 (100\%). Setting 01 increases from 50\% to 100\%, and the other eleven settings remain at 100\% in both versions. Main Table 1 summarizes the paired result; Figure~\ref{supp:fig:coverage} shows the full layout and object-parameter coverage.

\subsection{Three perception-conditioned whole-body
sequences}\label{three-perception-conditioned-whole-body-sequences}

Three whole-body conditions use measured (64\ensuremath{\times}227 mm, 600 g), narrow (58\ensuremath{\times}220 mm, 600 g), and wide-empty (68\ensuremath{\times}235 mm, 20 g) bottles. They complete eight-stage perception-conditioned sequences in \textbf{108, 112, and 108 s}, respectively, including bottle estimation, loaded walking, release, and return to stand.

\subsection{Physical deployment and execution}\label{physical-deployment-and-source-videos}

At the office site, an eight-stage MuJoCo validation executes a \textbf{124-second} task, taking \textbf{171 seconds} of planning and validation wall-clock time and placing the bottle within \textbf{4.2 cm} of the bin center. A separate MuJoCo bottle-to-bin rollout lasts \textbf{112 seconds}. G1 completes the continuous grasp--carry--dispose task in \textbf{101.7543 seconds} without retry. Door opening and cart transport provide a second \textbf{98.41-second} long-horizon physical sequence.

\begin{table}[htbp]\centering\small
\caption{Bottle-to-bin simulation variation units and completion checks.}\label{supp:tab:variation}
\begin{tabularx}{\linewidth}{@{}p{25mm}XX@{}}\toprule
\rowcolor{pastelviolet}\textbf{Study} & \textbf{Variation and criterion} & \textbf{Observed result} \\ \midrule
Room layout & 11 named, 12 development-random, 8 post-freeze; full grasp, carry, bin release, return. & 31/31 full-task completions; 61.2--138.6 s. \\
Bottle properties & 58--68 mm diameter, 220--235 mm height, 20--600 g; hand fixture at 8--10 s. & 44/44 opposing-grasp windows. \\
Paired refinement & The same 12 LHS settings, earlier versus refined contact window. & 11/12 to 12/12 full-window retention. \\
Visual whole-body & Measured, narrow-heavy and wide-light parameter conditions. & 3/3 eight-stage completions. \\ \bottomrule
\end{tabularx}\end{table}

\section{Quantitative skill-development
histories}\label{quantitative-skill-development-histories}

\subsection{Cohort and measurements}\label{cohort-and-measurements}

The study includes 34 consecutively registered skills (Cases 40--73), with 33 task requests matched to first human-approval timestamps. Each skill is one observation. Agent-turn time is the union of open-turn intervals, including tool and resource waits; calendar time spans request to first approval, including gaps and human review. Distinct retained full-body trajectories define dynamic versions, and image/frame-review calls define visual inspections. Median first-approval budgets (IQR) are \textbf{97.0 min} (42.5--170.3) agent-turn time, \textbf{2.80 h} (1.33--9.80) calendar time, \textbf{2 turns} (1--2), \textbf{18 versions} (7--31), \textbf{10 visual inspections} (7--18), and \textbf{1 human revision prompt} (0--1). For 32 later presentation-aligned skills, cumulative medians are \textbf{123.3 agent-turn minutes} and \textbf{3.89 calendar hours}.

The empirical distributions show \textbf{25/33} first approvals within three agent-turn hours, \textbf{21/33} within 20 retained dynamic versions, and \textbf{17/33} within ten visual inspections.

\begin{figure}[htbp]
\centering
\includegraphics[width=\linewidth]{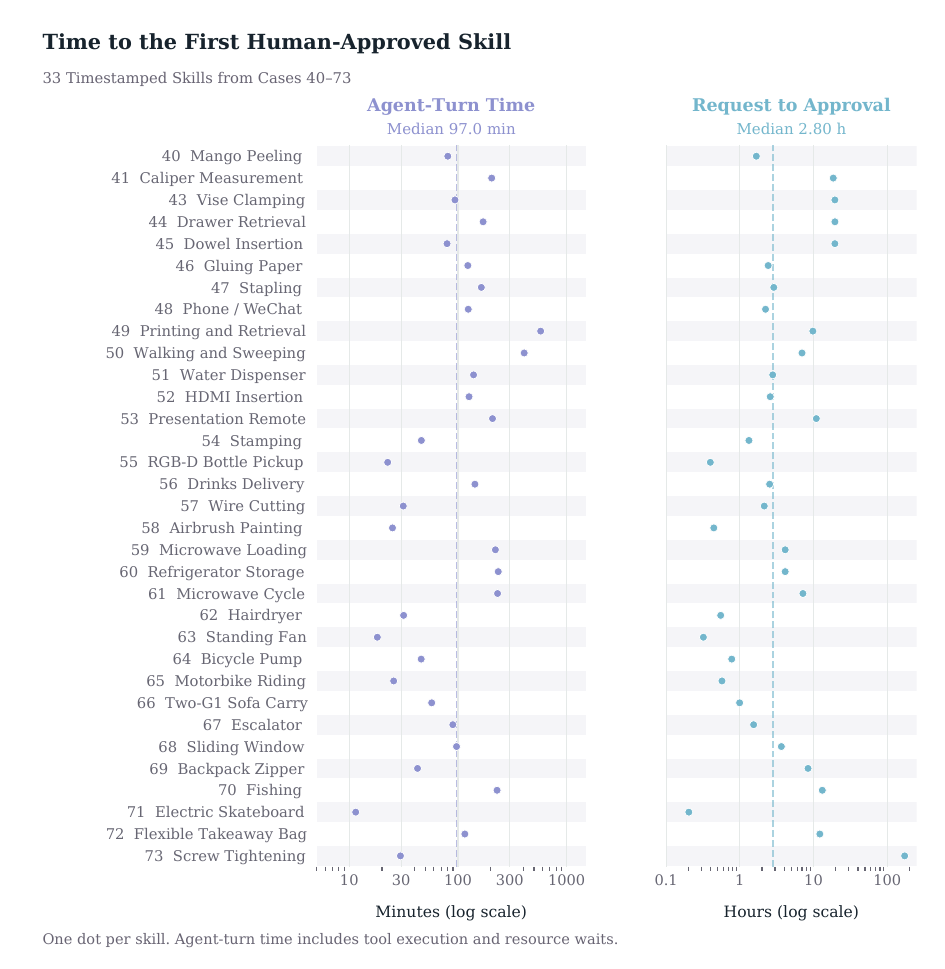}
\caption{Time to first human approval for 33 matched skills. Left: recorded agent-turn minutes; right: request-to-approval calendar hours on separate logarithmic axes.}
\label{supp:fig:time}
\end{figure}

\begingroup\small\renewcommand{\arraystretch}{0.94}
\begin{longtable}{@{}rp{42mm}rrrr@{}}\caption{First human-approval development records for 33 numbered skills. Agent time in minutes, calendar time in hours.}\label{supp:tab:skills}\\
\toprule\rowcolor{pastelviolet} ID & Skill & Agent & Calendar & Versions & Visual \\ \midrule\endfirsthead
\toprule ID & Skill & Agent & Calendar & Versions & Visual \\ \midrule\endhead
40 & Mango peeling & 80.6 & 1.67 & 9 & 17 \\
41 & Caliper measurement & 204.2 & 18.52 & 7 & 18 \\
43 & Vise clamping & 93.8 & 19.51 & 5 & 6 \\
44 & Drawer retrieval & 170.3 & 19.53 & 8 & 13 \\
45 & Dowel insertion & 79.3 & 19.48 & 3 & 9 \\
46 & Gluing paper & 123.2 & 2.42 & 31 & 20 \\
47 & Stapling & 164.1 & 2.89 & 18 & 8 \\
48 & Phone/WeChat & 124.3 & 2.23 & 12 & 24 \\
49 & Printing/retrieval & 576.1 & 9.80 & 58 & 16 \\
50 & Walking/sweeping & 406.5 & 6.98 & 47 & 29 \\
51 & Water dispenser & 139.1 & 2.79 & 18 & 17 \\
52 & HDMI insertion & 126.4 & 2.58 & 19 & 15 \\
53 & Presentation remote & 207.8 & 10.92 & 19 & 13 \\
54 & Stamping & 46.1 & 1.33 & 13 & 7 \\
55 & RGB-D bottle pickup & 22.6 & 0.40 & 6 & 3 \\
56 & Drinks delivery & 143.2 & 2.54 & 46 & 19 \\
57 & Wire cutting & 31.4 & 2.15 & 8 & 8 \\
58 & Airbrush painting & 25.0 & 0.44 & 4 & 7 \\
59 & Microwave loading & 221.0 & 4.13 & 34 & 18 \\
60 & Refrigerator storage & 234.6 & 4.12 & 64 & 10 \\
61 & Microwave cycle & 231.3 & 7.18 & 32 & 22 \\
62 & Hairdryer & 31.6 & 0.55 & 7 & 10 \\
63 & Standing fan & 18.1 & 0.32 & 5 & 5 \\
64 & Bicycle pump & 45.9 & 0.78 & 19 & 5 \\
65 & Motorbike riding & 25.6 & 0.57 & 16 & 8 \\
66 & Two-G1 sofa carry & 57.3 & 0.99 & 22 & 8 \\
67 & Escalator & 89.7 & 1.54 & 21 & 13 \\
68 & Sliding window & 97.0 & 3.66 & 24 & 9 \\
69 & Backpack zipper & 42.5 & 8.43 & 17 & 7 \\
70 & Fishing & 228.8 & 13.16 & 48 & 26 \\
71 & Electric skateboard & 11.5 & 0.20 & 6 & 6 \\
72 & Flexible takeaway bag & 115.7 & 12.18 & 34 & 19 \\
73 & Screw tightening & 29.6 & 172.42 & 5 & 7 \\\end{longtable}\endgroup

\begin{figure}[htbp]
\centering
\includegraphics[width=\linewidth]{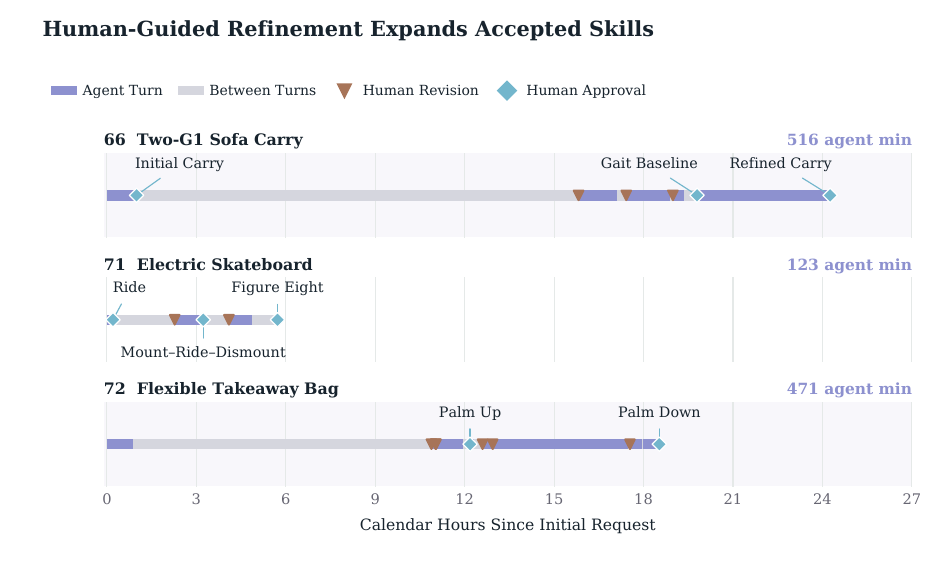}
\caption{Human-guided refinement and approval milestones for Cases 66, 71 and 72: two-humanoid sofa transport, electric skateboard and flexible-bag manipulation.}
\label{supp:fig:timeline}
\end{figure}

\subsection{Screw-tightening candidate
outcomes}\label{screw-tightening-candidate-outcomes}

\begin{figure}[htbp]
\centering
\includegraphics[width=\linewidth]{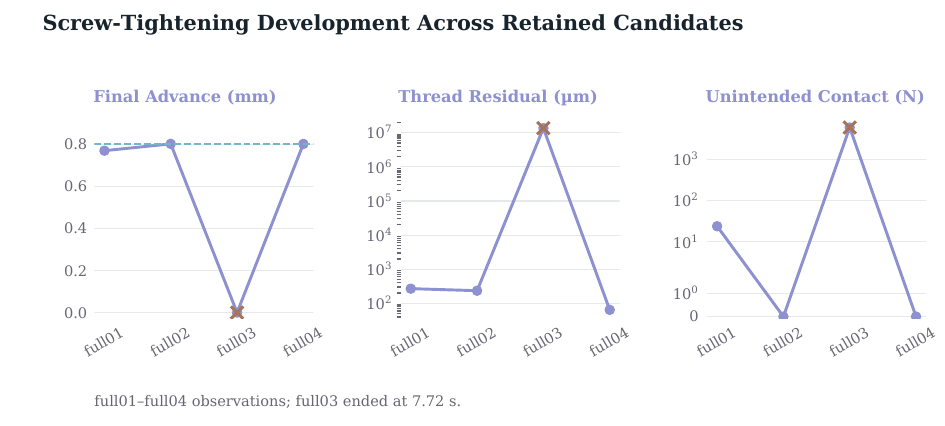}
\caption{Complete measured screw-tightening candidate sequence full01--full04, including the early-ended full03 trace.}
\label{supp:fig:screw}
\end{figure}

Screw-tightening development reduces maximum thread residual from \textbf{269.44 to 64.58 \ensuremath{\mu}m} (\textbf{76.0\%}) and moves final advance from \textbf{0.7697 to 0.8019 mm} against a \textbf{0.8 mm} target. Peak unintended contact decreases from \textbf{23.95 N to 0 N}. Figure~\ref{supp:fig:screw} presents the \texttt{full01}--\texttt{full04} candidate sequence, including an early-ended trajectory.

\section{BC/DAgger training and execution
interface}\label{bcdagger-training-and-execution-interface}

\subsection{Input and reference
target}\label{input-and-reference-target}

The shared policy has \textbf{3,127,287 parameters} and predicts six references from $t$ to $t+0.10$ s at \textbf{20-ms} spacing. Each 59-dimensional output comprises \textbf{29 body-joint targets}, \textbf{12 hand-joint targets}, \textbf{six palm position components} (three per palm), and \textbf{12 palm 6D rotation components}. Inputs comprise \textbf{176 task/robot observation slots} with skill and role indicators (20 and five categories), task phase, public clock, root/palm state, and explicit geometry memory, plus \textbf{32 mechanism values} and corresponding validity-mask bits. The reference student uses the current phase and geometry memory (\texttt{reference\_history=False}); the frozen whole-body tracker maintains its native proprioceptive history.

At 50 Hz, the execution loop decodes the six-frame predictions into body, hand, and palm references and interpolates between samples. Frozen ScaleBFM tracks the body, while bounded actual-state IK and hand/contact feedback adjust targets according to task stage. Object-frame codecs and event gates are shared with the program executor.

\begin{table}[htbp]\centering\small
\caption{Final BC/DAgger student observation and reference interface.}\label{supp:tab:dagger}
\begin{tabularx}{\linewidth}{@{}p{27mm}X@{}}\toprule
\rowcolor{pastelviolet} Component & Shared policy configuration \\ \midrule
Observation & 176 task/robot state slots, 32 mechanism values with 32 validity-mask bits, skill and phase identifiers, public clock and explicit geometry memory. \\
History & \texttt{reference\_history=False}; current public phase and geometry memory condition reference prediction. \\
Output & Six references $t,t+0.02,\ldots,t+0.10$ s, each 59 channels: 29 body, 12 hand, 6 palm positions and 12 palm 6D rotations. \\
Executor & 50 Hz reference decoder, frozen ScaleBFM tracker, bounded arm IK and task-event/contact feedback. \\
Training & BC initialization, student-state DAgger expert labeling and event-gated recovery corrections. \\ \bottomrule\end{tabularx}\end{table}

\subsection{Expert labeling, recovery and observed training
stages}\label{expert-labeling-recovery-and-observed-training-stages}

BC trains the reference student on targets from agent-authored skill programs and simulated executions. DAgger adds expert labels at student-visited states. Task-event gates can hold a phase or request a new reference when the expected physical transition has not occurred, as in stapling, caliper contact, and scissor cutting. Recovery labels encode these state-dependent corrections.

The ten-task extension dataset comprises \textbf{63 episodes and 92,130 frames}, including \textbf{77,130 student-execution frames} and \textbf{142 recovery-flag frames} within the same collection. Across joint system-development checkpoints, completion progresses from BC 2/10 through D1 5/10, D2 7/10, D3--D4 8/10, D6 9/10, and D7 10/10. These checkpoints integrate DAgger data aggregation with reference-prediction, representation, and execution-interface improvements.

\subsection{Fixed suite and task
galleries}\label{fixed-suite-and-task-galleries}

One shared checkpoint completes \textbf{18/18 simulated development tasks} and \textbf{18/18 same-condition repetitions} with \textbf{zero inference-time expert queries or takeovers}. The development rollouts total \textbf{673 simulated seconds}, with individual tasks lasting \textbf{16--66 s}. Frozen ScaleBFM and the task-feedback executor remain shared across the task suite.

\begin{table}[htbp]\centering\small\caption{Eighteen simulated tasks under the same shared reference checkpoint.}\label{supp:tab:policy}\begin{tabularx}{\linewidth}{@{}Xrrr@{}}\toprule\rowcolor{pastelviolet} Task & Time (s) & Dev. & Repeat \\ \midrule
Bottle pickup & 60 & yes & yes \\
Light switch & 48 & yes & yes \\
Drawer & 66 & yes & yes \\
Stapler & 31 & yes & yes \\
Standing fan & 50 & yes & yes \\
Faucet & 28 & yes & yes \\
Vise clamping & 48 & yes & yes \\
Drill & 42 & yes & yes \\
Hairdryer & 36 & yes & yes \\
Scissors & 24 & yes & yes \\
Wire cutting & 16 & yes & yes \\
Bakery tongs & 34 & yes & yes \\
Mango peeling & 30 & yes & yes \\
Caliper & 24 & yes & yes \\
Key & 26 & yes & yes \\
Tweezers & 26 & yes & yes \\
Umbrella & 34 & yes & yes \\
Saw & 50 & yes & yes \\\bottomrule\end{tabularx}\end{table}

\begin{figure}[htbp]
\centering
\includegraphics[width=\linewidth]{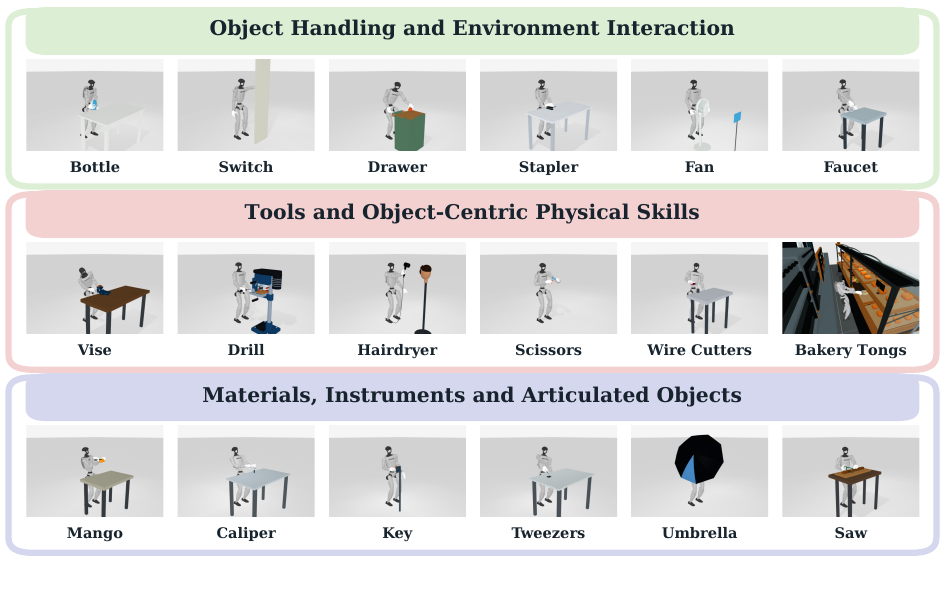}
\caption{Eighteen simulated tasks executed under one shared reference-policy checkpoint.}
\label{supp:fig:gallery}
\end{figure}

\begin{figure}[htbp]
\centering
\includegraphics[width=\linewidth]{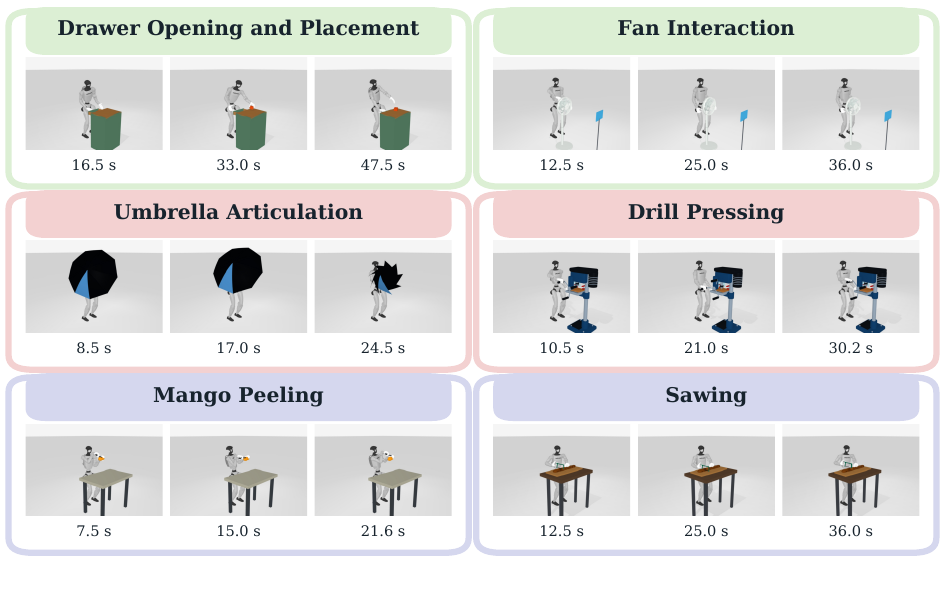}
\caption{Six time-ordered task sequences under the same shared reference checkpoint.}
\label{supp:fig:strip}
\end{figure}

The same reference model executes the full fixed task suite.

\section{Physical demo registry and reproducibility-facing
materials}\label{physical-demo-registry-and-reproducibility-facing-materials}

Six physical G1 demonstrations span five interaction categories: bottle disposal, door/cart transport, chair sit-to-stand, wall-switch actuation, and standing mouse drawing. Table~\ref{supp:tab:demos} reports sequence durations and task phases.

\begin{table}[htbp]\centering\small\caption{Six G1 demonstrations spanning five physical task categories.}\label{supp:tab:demos}
\begin{tabularx}{\linewidth}{@{}p{29mm}p{23mm}X@{}}\toprule\rowcolor{pastelviolet} Task & Duration & Physical sequence \\ \midrule
Bottle disposal & 101.7543 s & Grasp, turn, carry, bin release, retreat. \\
Door and cart transport & 98.41 s & Door opening and pushing a cart through doorway. \\
Chair sit-to-stand (unsupported) & 24.74 s & Sit, hold and return to stand. \\
Chair sit-to-stand (assisted) & 24.99 s & Sit, hold and return to stand. \\
Light switch & 20.48 s & Wall-switch actuation. \\
Standing mouse drawing & 28.26 s & Desktop drawing with physical mouse. \\ \bottomrule\end{tabularx}\end{table}